\documentclass[journal]{IEEEtran}
\usepackage{amsmath,graphicx}
\usepackage{amssymb,amsfonts}
\usepackage{algorithmic}
\usepackage{graphicx}
\usepackage{textcomp}
\usepackage{mathtools}
\usepackage{fixmath}
\usepackage[normalem]{ulem}
\usepackage{multirow}
\usepackage[dvipsnames,table,xcdraw]{xcolor}
\usepackage{hyperref}
\usepackage{cite}
\usepackage{soul}
\newcommand{\s}{\textcolor{black}} 
\newcommand{\okk}{\textcolor{black}}

\newcommand{\shr}{\textcolor{black}} 

\ifCLASSINFOpdf
\else
\fi
\begin{document}
%\title{Syllable Count Estimation using WaveCount}
\title{SylNet: An Adaptable End-to-End Syllable Count Estimator for Speech}

\author{Shreyas~Seshadri~and~Okko~R{\"a}s{\"a}nen%
\thanks{Shreyas Seshadri is with the Department
of Signal Processing and Acoustics, Aalto University, Finland. e-mail: (shreyas.seshadri@aalto.fi)}% <-this % stops a space
\thanks{Okko~R{\"a}s{\"a}nen is with the Faculty of Information Technology and Communication Sciences,  Tampere University, Finland, and the Department of Signal Processing and Acoustics, Aalto University, Finland.}}

%\markboth{Journal of \LaTeX\ Class Files,~Vol.~14, No.~8, August~2015}%
%{Shell \MakeLowercase{\textit{et al.}}: Bare Demo of IEEEtran.cls for IEEE Journals}

\maketitle

\begin{abstract}
Automatic syllable count estimation (SCE) is used in a variety of applications ranging from speaking rate estimation to detecting social activity from wearable microphones or developmental research concerned with quantifying speech heard by language-learning children in different environments. The majority of previously utilized SCE methods have relied on heuristic DSP methods, and only a small number of bi-directional long short-term memory (BLSTM) approaches have made use of modern machine learning approaches in the SCE task. This paper presents a novel end-to-end method called SylNet for automatic syllable counting from speech, built on the basis of a recent developments in neural network architectures. We describe how the entire model can be optimized directly to minimize SCE error on the training data without annotations aligned at the syllable level, and how it can be adapted to new languages using limited speech data with known syllable counts. Experiments on several different languages reveal that SylNet generalizes to languages beyond its training data and further improves with adaptation. It also outperforms several previously proposed methods for syllabification, including end-to-end BLSTMs. 
\end{abstract}

% Note that keywords are not normally used for peerreview papers.

%\IEEEpeerreviewmaketitle
\section{Introduction}\label{sec:intro}

Automatic syllable count estimation (SCE) from speech is used in a number of applications. For instance, syllable-based speaking rate estimation algorithms (such as \cite{emrate, wang2007robust}) can be used to analyze prosodic patterns of speakers and speaking styles for linguistic research, or used as additional 
information for training text-to-speech (TTS) synthesis systems. Syllables are also used for automatic estimation of vocal activity and social interaction from long and noisy audio recordings captured by wearable microphones, as in the personal life log application of Ziaei et al. \cite{ziaei2015}\cite{ziaei2016}. There is also a need for robust language-independent methods for quantifying the amount of speech in daylong child-centered audio recordings from various language environments \cite{ACLEW, RasanenWCE}, as child language researchers use such data to understand language development in children in  (e.g., \cite{Bergelson2018, Weisleder2013}). 

The main motivation for choosing SCE over automatic speech recognition (ASR) in these tasks comes from the robustness and potential language-independence of the syllabification algorithms. While ASR may be applicable in high-resource domains and good recording conditions, deployment of reliable ASR systems across a large set of (potentially low-resource) languages is still challenging in a manner that also ensures consistent performance in the adverse signal conditions typical to daylong recordings from wearable devices. In contrast, syllables are fundamentally based on the rhythmic succession of low- and high sonority speech sounds present in all languages. Since sonority is closely related to the amplitude modulation of speech signals, with syllabic nuclei being the locally most energetic speech sounds  (see, e.g.,\cite{parker2002} or \cite{rasanen2018pre} for an overview), syllables are potentially robust against noisy environments and channel variations.

However, so-far the most widely used syllabifiers (segment boundary and nuclei detectors) have been based on heuristic DSP operations \cite{emrate}\cite{syllomatic} \cite{wang2007robust}\cite{rasanen2018pre}\cite{mermelstein}\cite{villing}\cite{reddy2013}. According to our knowledge, only the bi-directional long short-term memory (BLSTM) -based algorithms by \cite{landsiedel2011syllabification, RasanenWCE}, and \cite{jiao2016} have used modern machine learning techniques to solve the SCE problem. In addition, the actual cross-language generalization capabilities of the algorithms, including \cite{landsiedel2011syllabification} and \cite{jiao2016}, have largely remained untested, nearly all studies using speech data from only one language for development and testing. 

This paper addresses the gaps in the literature by proposing a new method called SylNet for automatic SCE from speech, making use of state-of-the-art WaveNet-like \cite{vanwavenet} architecture for end-to-end SCE. We benchmark SylNet and a number of existing unsupervised and supervised algorithms on three languages different from the training data. We also investigate adaptability of different approaches for improved performance when a small amount of speech with known syllable counts is available in the target language. Unlike the previous BLSTM-models by \cite{landsiedel2011syllabification} and \cite{RasanenWCE} and similar to \cite{jiao2016}, the newly proposed SylNet does not require temporally aligned syllable- and phone-level annotations for training or adaptation. As a result, we show that SylNet achieves state-of-the-art SCE performance on novel languages without any language-specific tuning, and can be improved even further with adaptation.

\section{SylNet}\label{sec:waveCount}
\begin{figure}[t!]
    \centering
    \vspace{-1.5mm}
    \includegraphics[width=1\linewidth]{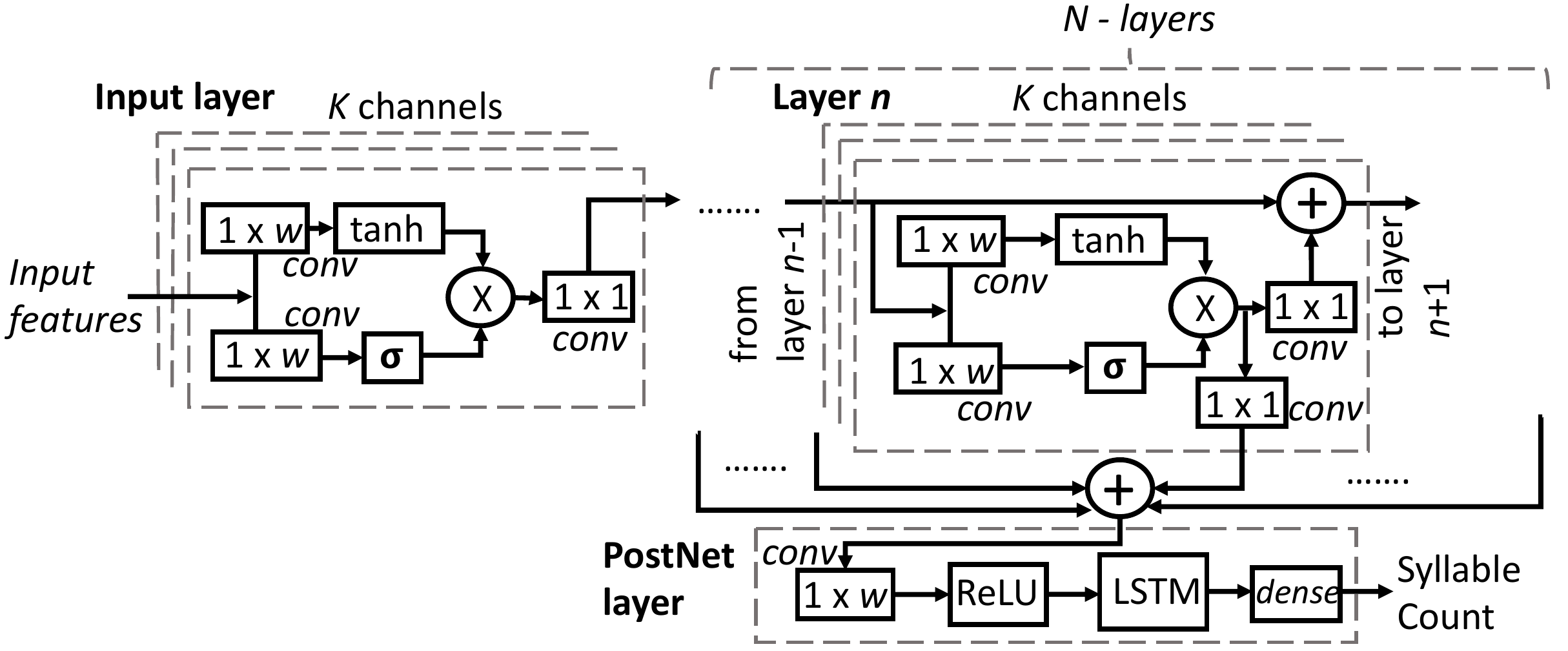}
    \vspace{-6.5mm}
    \caption{Structure of the SylNet model with input layer on the left and the rest of the $N$ layers on the right. Each of the $N$ layers consists of $K$ channels, all feeding a shared PostNet layer that integrates information across the entire input using an LSTM, and then to a dense layer that maps activations from all scales and channels into one scalar number (when using L1 loss in Eq. \ref{eq:rel_err}) or vector (ordinal loss in Eq. \ref{eq:ord_err}) corresponding to estimated syllable count.}
    \vspace{-4.5mm}
    \label{fig:block}
\end{figure}

Figure \ref{fig:block} shows the structure of the SylNet model\footnote{Python + TensorFlow implementation of SylNet with a pre-trained language-independent model is available for download at \url{https://github.com/shreyas253/SylNet}.}, which is derived from the recently developed WaveNet \cite{vanwavenet} model used for speech synthesis. Instead of autoregressive waveform sample prediction as in speech synthesis, we apply the model, along with an additional LSTM layer, to the prediction of accumulated syllable counts based on spectral features of speech. The motivation for using a WaveNet-like architecture lies in the efficient training and use of multi-scale information in speech signals and easy adaptability of the network through tuning only the parameters responsible for merging information from different feature extraction layers.

Input to the model consists of time-frequency features of speech sampled at a regular interval ($T$ timesteps, $D$ dimensions). The input layer has $K$-channels each consisting of a $w$-point gated convolutional unit. This is followed by $N$ layers, each consisting of similar gated convolutional units as the input layer. The outputs of these gated convolutional units are passed through an affine transform and added with a residual \cite{resnet2016} of the original input, and then passed to the next layer. 
%The residual connection helps in preventing the problem of diminishing gradients in deeper networks, \s{by allowing the error gradients to pass through} \cite{resnet2016}. 
Skip connections\cite{long2015fully} pass the outputs of the gated convolutional unit from each of the $N$ layers to a so-called PostNet layer through separate layer-specific affine transformations. The residual \cite{resnet2016} and skip \cite{long2015fully} connections help in preventing the problem of diminishing gradients in deeper networks by allowing the error gradients to pass through, thus reducing convergence time and enabling the use of deep models \cite{vanwavenet}. The PostNet layer integrates the information coming from the skip connections of all the $N$ layers through addition. This sum is then fed to a rectified linear unit (ReLU) % non-linearity 
after a $w$-point affine convolution, %finally, 
to an LSTM layer that accumulates syllable count information over time%, \shr{(see L1 loss, in Eq. \ref{eq:rel_err} and ordinal loss in Eq. \ref{eq:ord_err}).}
, and finally to a dense layer with 1 or R units (depending on the loss; Eqs. \ref{eq:rel_err} and \ref{eq:ord_err}).   %((in case of L1 loss, in Eq. \ref{eq:rel_err}) or R units %(Eq. \ref{eq:ord_err}).

During training, only the PostNet outputs of the last time frame for each utterance are used in the calculation of the loss function, which compares estimated syllable counts to the known ground truth for that utterance (Eqs. \ref{eq:rel_err} and \ref{eq:ord_err}).
As in any deep convolutional architecture, the $N$ hidden layers can be considered as feature extractors that compute time-frequency patterns of an increasing non-linearity and complexity that are relevant for the current task. The PostNet layer does the job of assimilating these features to the syllable count. This structure has the advantage that a pre-trained model can be adapted to another (potentially much smaller) dataset in different language by only re-training the PostNet layers that only have a limited number of parameters. As we show in the experiments, this avoids overfitting even when very little adaptation data are available.

\section{Compared SCE Methods}\label{sec:methods}
Our experiments in Section V compare the SylNet to a number of alternatives from two classes of SCE algorithms: so-called envelope-based methods and end-to-end methods, where the latter category also includes the present SylNet approach. 
%The methods studied here can be categorised as follows
\subsection{Envelope prediction approaches}
\begin{figure}[h]
    \centering
    \vspace{-3.5mm}
    \includegraphics[width=0.95\linewidth]{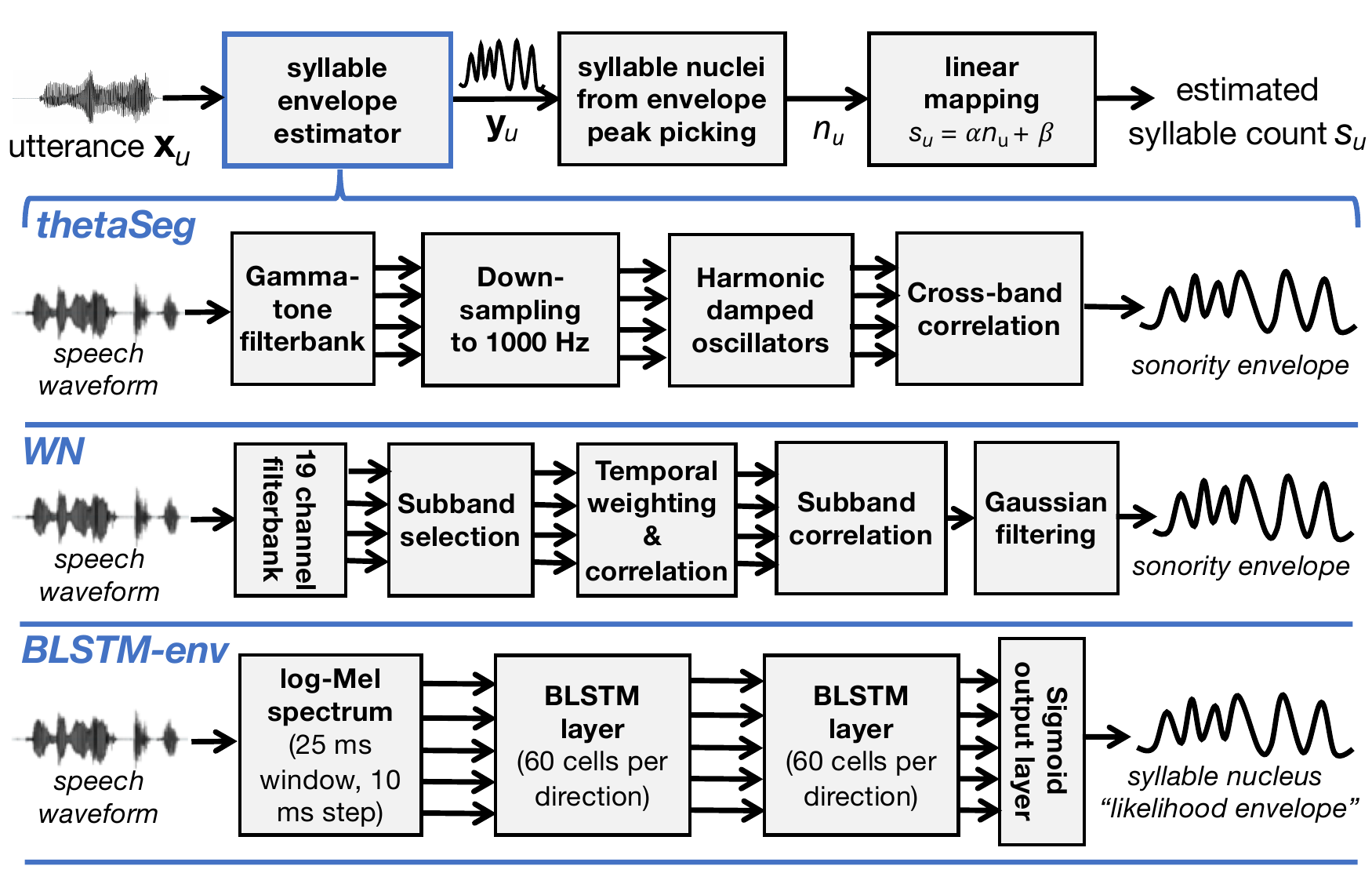}
    \vspace{-5.5mm}
    \caption{Block diagrams depicting the basic structure of the envelope estimation methods used for syllable count estimation (adapted from \cite{rasanen2018comparison}).} 
    \vspace{-4mm}
    \label{fig:block_2}
\end{figure}

All envelope-based SCE methods consist of two basic stages (see Fig. \ref{fig:block_2}): 1) computation of a so-called sonority envelope for the speech input, in which temporally local peaks are assumed to correspond to syllabic nuclei in the input, and 2) automatic peak picking to extract nuclei count from the envelope. Optionally, a third stage consisting of a least-squares linear model from model outputs to reference syllable counts (or to speaking rate as in \cite{wang2007robust}, or word counts, as in \cite{ziaei2016, RasanenWCE, rasanen2018comparison}) can be used, allowing correction of any systematic biases in the estimation procedure. The envelope estimation can be either based on a series of DSP operations (hence called a "blind" or "unsupervised" approach), or by a supervised machine-learning approach to map incoming speech into syllabic nuclei likelihoods.
%(with a threshold that determines the minimum amplitude difference between a local maximum and the previous local minimum in order for the maximum to be considered as a nucleus), 

In this study, two previously reported unsupervised methods were compared: "\textbf{thetaSeg}" that carries out perceptually motivated entrainment to sonority fluctuations in speech with proven cross-language capability for syllable segmentation \cite{rasanen2018pre}, and the envelope detector stage from speaking rate estimator by Wang and Narayanan ("\textbf{WN}", \cite{wang2007robust}) that was also found to be the best syllabifier for word count estimation in \cite{ziaei2016}. In addition, we include a supervised BLSTM-based system described in \cite{rasanen2018comparison}, developed further from the one reported in \cite{landsiedel2011syllabification}, here referred to as \textbf{BLSTM-env}. The first two consist of a series of DSP operations to transform signal waveform \textbf{x}$_u$ of an utterance $u$ into a smooth sonority envelope \textbf{y}$_u$ to be used for peak picking. In contrast, BLSTM-env consists of two stacked BLSTM layers, each having 60 LSTM cells, that are responsible for converting 24-channel log-Mel (25-ms window with 10-ms time steps) input features into syllable nucleus likelihoods \textbf{y}$_u$ as a function of time, again followed by peak picking. In BLSTM-env training, nucleus targets are presented as Gaussian-shaped kernels centered around the nucleus phone of each manually annotated syllable (see \cite{landsiedel2011syllabification}). The basic structure of these methods is elaborated in Figure \ref{fig:block_2}, and more details can be found in the respective papers (\cite{rasanen2018pre}\cite{wang2007robust},\cite{rasanen2018comparison}).

For all three envelope-based approaches, the initial syllable counts $n${$_u$} are acquired with a peak picking algorithm that uses a threshold hyperparameter $\theta$, which determines the minimum required amplitude difference between a local maximum and the previous local minimum in order for the maximum to be considered as a nucleus. These counts are then mapped to final syllable count estimates using a linear mapping $s$$_u$ = $\alpha n$$_u$+$\beta$. During training $\theta$, $\alpha$, and $\beta$ are jointly estimated using an exhaustive search over all possible thresholds $\theta$ with their corresponding least-squares linear models, choosing the triplet that the minimizes relative error in Eq. (1) on the training data. We did not optimize the large number of internal parameters (filter lengths etc.) of the unsupervised methods, but used the ones found to be optimal in the original papers \cite{wang2007robust}\cite{rasanen2018pre}.

\subsection{End-to-end methods}\label{sec:dcmethods}
\begin{figure}[h]
    \centering
    \vspace{-4mm}
    \includegraphics[width=0.95\linewidth]{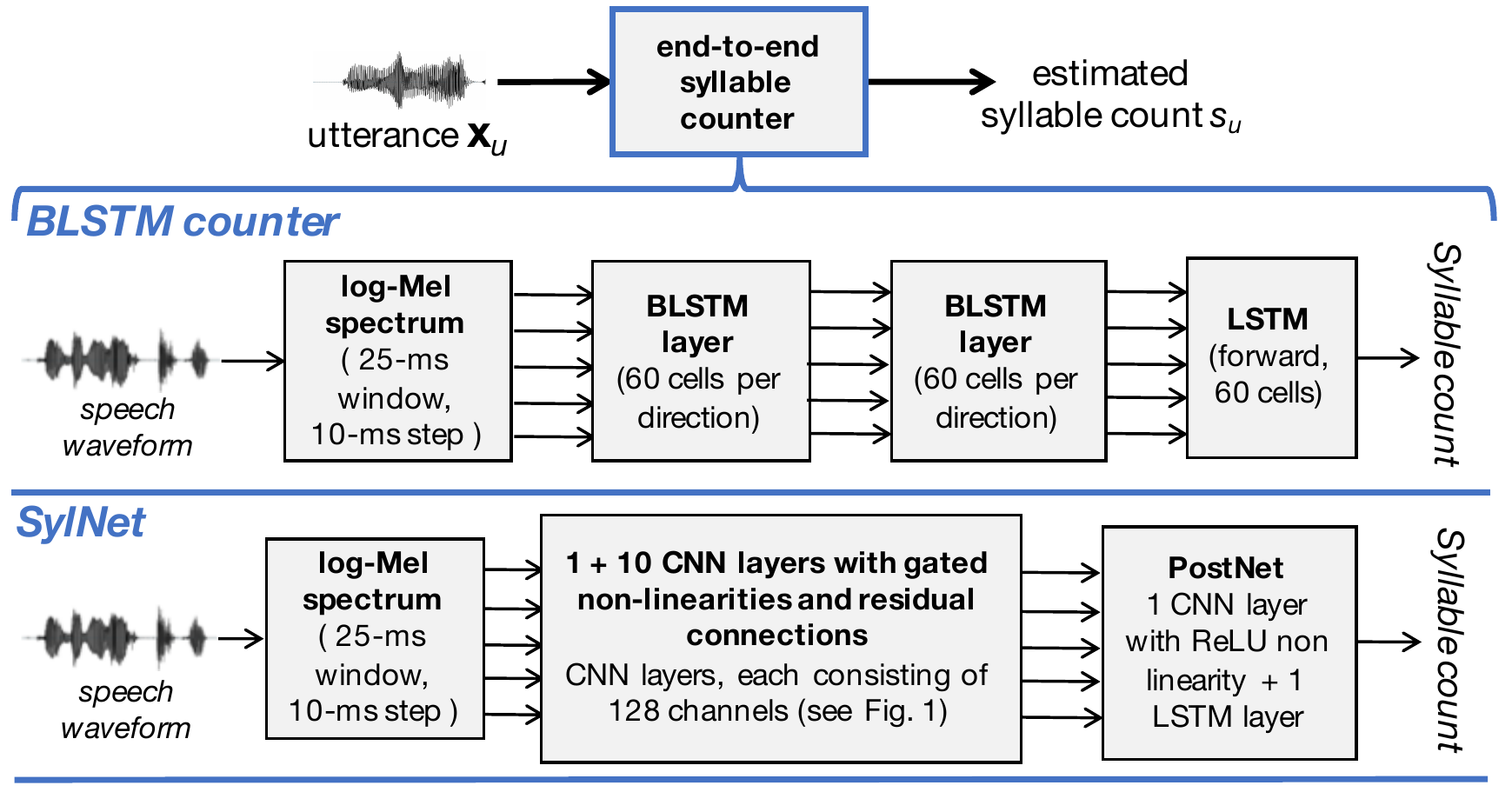}
    \vspace{-3mm}
    \caption{Block diagrams depicting the basic structures of the end-to-end SCE methods for syllable count estimation.} 
    \vspace{-3mm}
    \label{fig:block_3}
\end{figure}

End-to-end methods aim to directly predict the syllable count from speech utterances without any hand-specified intermediate representational stages. We examine two supervised neural network methods for end-to-end SCE that both use 24-channel log-Mel speech features 
%(25-ms window with 10-ms time steps) 
as inputs (as in the BLSTM-env; see Fig. \ref{fig:block_3}):
1) \textbf{BLSTM-count}, which is a BLSTM network having five hidden LSTM layers (2 stacked standard BLSTMs + a forward LSTM layer with linear activation), i.e., a similar structure to the BLSTM-env, but now having an additional forward LSTM layer to output estimated syllable counts at the end of the input utterance (see also \cite{jiao2016}). Each layer has 60 LSTM cells (per direction).

2) The newly proposed \textbf{SylNet}. After initial experimentation on training data (see Section IV), $N$ and $K$ were set to 10 and 128, respectively. Convolution kernel length $w$ was set to 5, with the effective receptive field of the topmost layer corresponding to 53 frames. The final LSTM layer in the PostNet was also set to have 128 cells.
The basic structure of the model was described in Section II, and is elaborated in Figures \ref{fig:block} and \ref{fig:block_3}.

\textit{Training Loss - }We explored two different loss functions and associated output layers for SylNet training. The first loss term (also used by the BLSTM-count) was defined as the relative error between true and estimated syllable counts, as this metric was also used as the evaluation metric in the experiments due to its previous use in the evaluation of word-count estimation systems \cite{ziaei2016, RasanenWCE, rasanen2018comparison}. L1-based relative error was defined as:
\begin{equation}
    \label{eq:rel_err}
    \begin{aligned}
        e_{L1} = \frac{1}{M}\sum_{u=1}^{M}\frac{\mid \widehat{s_u}-s_u \mid}{s_u}
    \end{aligned}
\end{equation}
where $\widehat{s_u}$ and $s_u$ correspond to the predicted and true target number of syllables for the $M$ training utterances in each mini-batch respectively. With this loss, last layer of the network was single linear unit performing regression to $\widehat{s_u}$. 

 However, the true syllable counts are discrete and ordinal in nature. %The simplest method to handle ordinal data is to treat the targets as continuous real values (see \cite{gutierrez2016ordinal} for a summary of handling ordinal data). 
Another potential loss function, earlier found to be more suited to ordinal data \cite{cheng2008neural,niu2016ordinal,gutierrez2016ordinal}, treats the problem as a series of $R-1$ binary classifications with $R$ being the rank of the ordinal targets \shr{(set to the maximum number of syllables per utterance in the training data)}. Here the last layer of the DNN is defined as a set of $R-1$ sigmoid units. The $r$th output represents the probability that the value of the output is larger than $r$. For example, target count of three would be represented as [1 1 1 0 ... 0] with $R-1-3$ zeros. Based on initial experiments, the relative ordinal objective loss between the sigmoid \shr{activations} $\widehat{o_u}$ and %\shr{the expected true targets of the sigmoid activations}
\okk{and the true sigmoid targets $o_u$ corresponding to the given reference count $s_u$}
%($o_u$) 
was defined as:
\begin{equation}
    \label{eq:ord_err}
    \begin{aligned}
        e_{ord} = \frac{1}{M}\sum_{u=1}^{M}\frac{\sqrt{ \widehat{o_u}^2-o_u^2}}{s_u}
    \end{aligned}
\end{equation}
and this was compared to the L1 loss in the experiments. For testing, $\widehat{s_u}$ is obtained by thresholding sigmoid outputs at 0.5.
%During testing the sigmoid outputs are thresholded to 0.5 (THIS LINE CAN BE SKIPPED).}

\section{Experimental Setup}
Our experimental setup aimed to test two conditions: 1) out-of-the-box generalization of the compared SCE algorithms to novel languages beyond the training data, and 2) performance on these novel languages when a limited amount of speech data with syllable counts (but not necessarily syllable positions) is available for algorithm adaptation.

\subsection{Data}
Two large speech corpora with syllable- and phone-level annotations were used in the for the initial training of the supervised syllable count estimation models and for optimization of detection thresholds and linear mapping coefficients of unsupervised methods: \textbf{1) Phonetic Corpus of Estonian Spontaneous Speech} (“EstPhon”; \cite{lippus2013phonetic}) and \textbf{2) Korean Corpus of Spontaneous Speech} (“Korean”; \cite{yun2015korean}). The former includes several spontaneous Estonian dialogues with pairs of male and female talkers, totaling up to 10,158 utterances in high studio quality (5.2 hours of audio). Korean corpus also consists of conversational dialogues between talkers, including a wide range of speaker ages. In order to keep the Korean corpus comparable to EstPhon in size, a randomly sampled a subset of 12,000 utterances (5.0 hours) was used for syllabifier training.

Three additional corpora with manually annotated syllable counts were chosen for testing and adaptation:
1) \textbf{FinDialogue Corpus of Spontaneous Finnish Speech} \cite{lennes2009segmental}, having dialogues from 4 talkers (2 male; 64 minutes of speech in total), 2) \textbf{C-PROM corpus of spoken French} \cite{avanzi2010c} consisting of 24 multi- minute recordings of various regional varieties of French from several discourse genres, totaling 1.2 hours of data, and 3) \textbf{Switchboard  corpus of spontaneous American English telephone conversations} (“SWB”; \cite{godfrey1992switchboard}), using a syllable-annotated ICSI transcription project subset (hundreds of talkers, 159 minutes). While FinDiag is of high studio quality, C-PROM contains more varied recording conditions (e.g., including some signal clipping), whereas SWB consists of narrowband telephone landline recordings, hence causing a substantial channel mismatch between training and testing.

\subsection{Training, Adaptation and Evaluation}\label{ssec:tr}
SylNet and BLSTMs \s{(with approximately 3M and 100k parameters, respectively)} were trained on Korean and EstPhon with the ADAM optimizer \cite{kingma2014adam} with hyperparameters $\beta_1$ and $\beta_2$ set to 0.9 and 0.999 and learning rate set to $10^{-4}$, with a 50\% dropout applied after every non-linearity, and an early stopping of 10 epochs. Minibatch size was set to 32 utterances. 
%After experimenting with several training strategies, 
%
% of 100, using ADAM optimizer for BLSTM-env estimator (same hyperparameters as above but learning rate was set to $10^{-3}$) and RMSprop for BLSTM-count (default Keras hyperparameters), and using the same early stopping criterion as above.
%\s{The BLSTM-count and SylNet models contain approx 100K and 3M parameters respectively.}
The test corpora were divided into separate adaptation and testing sets. Each test set was defined by randomly sampling approx. 50\% of all utterances of the corpus for the set. To simulate varying amounts of adaptation data, 8 different-sized sets ranging from mere 30 seconds up to 45 minutes of speech were created from the remaining utterances, ensuring that no same speaker occurred in both adaptation and testing. Five randomly sampled sets of each of size were created, and average performance across these are reported in the results.
For the envelope estimation methods, adaptation involved estimating the optimal peak picking  parameter $\theta_L$ and the linear model parameters $a_L$ and $b_L$ that minimize relative syllable count error on the full adaptation set $L$. 
%End-to-end methods were adapted by retraining the last LSTM layer \s{in the BLSTM-count and the PostNet in the WaveCount} until early stopping criterion on the validation split of the adaptation data was reached. 
BLSTM-count was adapted by retraining the last LSTM layer, and SylNet by retraining the entire PostNet module. 
%For the \s{SylNet}, the adaptation training was carried with more aggressive early stopping with of 3-5 epochs, a lower learning rate of $10^{-5}$-- $10^{-4}$ and a smaller minibatch size of 4--32 with smaller values for less data. BLSTM-count was re-trained until convergence with an early-stopping windows of 3 epochs and using a mini-batch size of 250 utterances, as this larger minibatch size during adaptation clearly improved the results from the 100 used for original model training. 
Similarly to \cite{ziaei2016, RasanenWCE, rasanen2018comparison}, the methods were evaluated on the held-out test data using the L1-relative error (\%) as shown in Eq. \ref{eq:rel_err}.

\section{Results}
\begin{figure}[t!]
    \centering
    \vspace{-1.5mm}
    \includegraphics[width=1\linewidth]{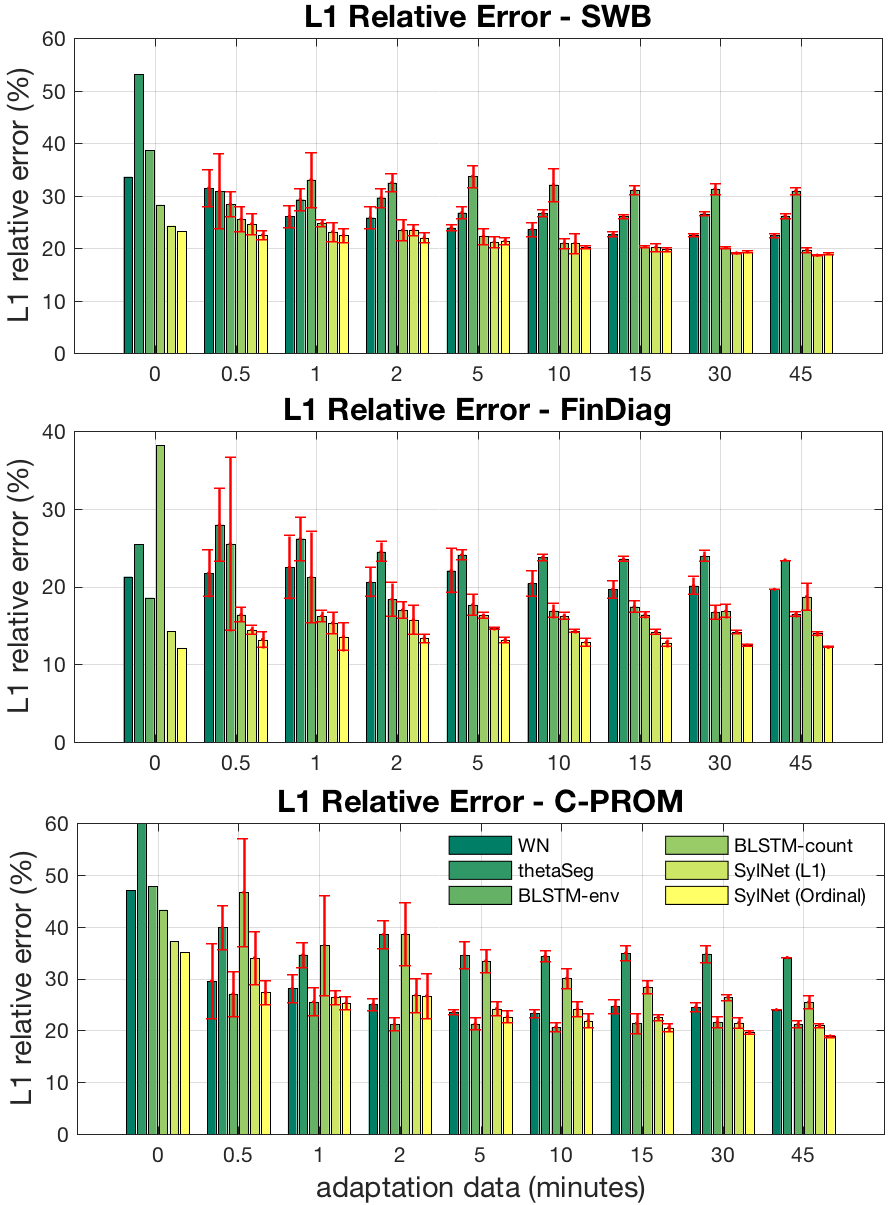}
    \vspace{-7.5mm}
    \caption{L1-Relative errors on the SWB, FinDiag and C-PROM datasets for difference syllable count estimation methods with varied amount of adaptation data. Error bars correspond to one standard deviation across the 5 randomly sampled folds for adaptation data of each amount (see Section \ref{ssec:tr}).}
    \vspace{-2mm}
    \label{fig:rel_err}
\end{figure}

\begin{figure}[h!]
    \centering
    \vspace{-1.5mm}
    \includegraphics[width=0.8\linewidth]{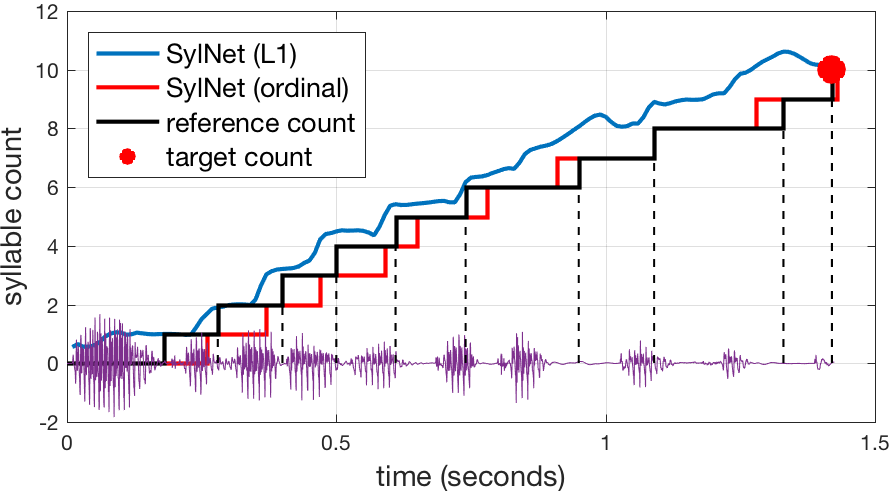}
    \vspace{-3.5mm}
    \caption{An example of SylNet PostNet output accumulation (using L1 and ordinal loss) as a function of time on a Finnish utterance from FinDiag.} %optimized with the L1 and Ordinal loss finctions (see Eq. \ref{eq:rel_err}, \ref{eq:ord_err}).}}
    %A causal version of the WaveNet, i.e., one that cannot see the future signal frames, is also shown for reference. }
    \vspace{-4.5mm}
    \label{fig:postnet_example}
\end{figure}

Figure \ref{fig:rel_err} shows the results in terms of L1 relative error. On average, the end-to-end methods outperform the envelope-based approaches, even though BLSTM-count fails on FinDiag before adaptation. All methods generally improve performance as the size of the adaptation dataset increases. Overall, the SylNet produces the lowest relative error before adaptation and after full adaptation, even though WN and BLSTM-env have comparable or sometimes better performance on some C-PROM adaptation sets. Also, the effects of end-to-end system adaptation on FinDiag dataset are minimal (approx. $\sim$2\% points for SylNet). This is likely because Finnish has relatively similar phonetic and syllabic structure to Estonian used for model training. As for the SylNet loss functions, the ordinal loss (Eq. \ref{eq:ord_err}) outperforms the L1 loss both out-of-the-box and after adaptation.

In order to visualize how the SylNet accumulates syllable counts over time, Fig. \ref{fig:postnet_example} shows the PostNet output activations for an utterance taken from the FinDiag corpus together with the accumulating reference syllable counts or both the L1 and the ordinal losses. Since receptive field size of the deepest convolutional kernels is very large in time, the model utilizes information from both past and future frames in order to optimize the count estimate at the final frame. In both cases for the utterance shown, the final syllable count estimate is very accurate.%If the model is restricted to past and present inputs (also shown in Fig. \ref{fig:postnet_example}), the PostNet accumulator follows the true syllable count more faithfully throughout the utterance. In both cases, the final syllable count estimate is very accurate.

\section{Conclusions}
This paper presented a novel end-to-end method called SylNet for syllable count estimation (SCE) from conversational speech and evaluated its performance on a number of languages, both out-of-the-box and by adapting the model to the target language. The results show that the proposed algorithm generalizes well to novel languages, and further improves with adaptation data if syllable counts (but not necessarily temporal alignments) are available for some minutes of speech in the target language. In total, the work shows that supervised end-to-end systems can be utilized for language-independent SCE from speech, clearly outperforming the earlier DSP-based methods. Future work should seek ways to further improve robustness and generalization capabilities of SylNet, e.g., through various data augmentation techniques. %The method should also be tested on daylong audio recordings used in application such as \cite{rasanen2018pre} and \cite{ziaei2016}) once datasets with gold standard syllable counts become available. 

\appendices
\section*{Acknowledgment}
This study was funded by Academy of Finland grants no. 312105 and 314602. The authors thank Lauri Juvela for his help with the SylNet implementation.

\newpage

{
\footnotesize
%\ninept

}
\bibliographystyle{IEEEtran}
\bibliography{ref}

\end{document}